\documentclass[twoside,leqno,twocolumn]{article}
\usepackage{ltexpprt}

\usepackage{amsmath,amssymb,latexsym,float,url,multirow,theorem,subfigure,epsfig,xspace,paralist,algorithm,algorithmic}
\newfloat{Algorithm}{thp}{}[section]
\newfloat{Algorithm}{tb}{lox}
\floatname{Algorithm}{Algorithm}
\begin{document}
\title{Graph Based Classification Methods Using Inaccurate External Classifier Information}
\author{S Sundararajan\thanks{Yahoo! Labs, Bangalore, India, ssrajan@yahoo-inc.com}\\
\and
S Sathiya Keerthi\thanks{Yahoo! Research, Santa Clara, CA, selvarak@yahoo-inc.com}}
\date{October 9, 2009}
\maketitle
\begin{abstract}\small\baselineskip=9pt
In this paper we consider the problem of collectively classifying entities where relational information is available across the entities. In practice inaccurate class distribution for each entity is often available from another (external) classifier. For example this distribution could come from a classifier built using content features or a simple dictionary. Given the relational and inaccurate external classifier information, we consider two graph based settings in which the problem of collective classification can be solved. In the first setting the class distribution is used to fix labels to a subset of nodes and the labels for the remaining nodes are obtained like in a transductive setting. In the other setting
the class distributions of {\it all} nodes are used to define
the {\it fitting} function part of a graph regularized objective function. We define a generalized objective function that handles both the settings. Methods like harmonic Gaussian field and local-global consistency (LGC) reported in the literature can be seen as special cases. We extend the LGC and weighted vote relational neighbor classification (WvRN) methods to support usage of external classifier information. We
also propose an efficient least squares regularization (LSR) based method and relate it to information regularization methods. All the methods are evaluated on several benchmark and real world datasets. Considering together speed, robustness and accuracy, experimental results indicate that the LSR and WvRN-extension methods perform better than other methods.
\end{abstract}



\section{Introduction}
Traditionally classifiers are built using only {\it local} features
of individual entities such as web pages or images. Relational classifiers also make use of relational information that exist across the entities. For example, local features of a web page could be collection of keywords that appear in the title or page content and useful relational information could be scores computed from presence/absence of inlinks and/or outlinks, similarity of page structure, url etc.
In relational classification problems, a collection of entities is represented as a graph or a network of nodes. Each node represents an entity with its set of local features, class attribute information and edge weight encodes any relational information that exists between those connected nodes. In its general setup the graph contains zero or more labeled nodes and one or more unlabeled nodes and the goal is to label a set of unlabeled nodes (test nodes). This problem is either solved by treating it as an induction problem where a model (possibly statistical) model is learnt using training data (labeled nodes) and used to classify future data, or, as a transduction problem where classification is needed(done) only on the test nodes. The key aspect of this relational classification problem is making {\it collective} inference of the class labels, that is, labels of all the nodes are obtained simultaneously. See \cite{priti}, \cite{sofus} and references there for more details.



In this paper we consider a related relational learning problem where, instead of a subset of labeled nodes, we have inaccurate external label/class distribution information for each node. This problem arises in many web applications. Consider, for example, the problem of identifying pages about Public works, Court, Health, Community development, Library etc. within the web site of a particular city. The link and directory relations contain useful signals for solving such a classification problem. Note that this relational structure will be different for different city web sites. If we are only interested in a small number of cities then we can afford to label a number of pages in each site and then apply transductive learning using the labeled nodes. But, if we want to do the classification on hundreds of thousands of city sites, labeling on all sites is expensive and we need to take a different approach. One possibility is to use a selected set of content dictionary features together with the labeling of a small random sample of pages from a number of sites to learn an inaccurate probabilistic classifier, e.g., logistic regression. Now, for any one city web site, the output of this initial classifier can be used to generate class distributions for the pages in the site, which can then be used together with the relational information in the site to get accurate classification.

The {\it problem of doing relational learning together with externally available class distribution information} can be solved by modifying existing transductive methods.
The problem has been discussed within broad sets of techniques such as {\it denoising}, {\it relaxation labeling}, {\it metric labeling} (see \cite{klein} and references there) etc., as well as within recent specific techniques such as the Gaussian Field Harmonic Function (GFHF) method \cite{zhu} and the Information Regularization (IR)\cite{cord}, dual IR (DIR)~\cite{koji} methods. We also consider two more methods, viz., the Local-Global Consistency (LGC) method \cite{zhou} and the probabilistic Weighted vote Relational Neighbor (WvRN) Classification method with Relaxation Labeling \cite{sofus}. The main aim of this paper is to take a few techniques (the last five specific methods mentioned above) that are popularly discussed and used in recent relational learning literature, extend them as needed and compare in different settings to see which ones are most effective in solving our problem. The proposed extensions and method include supporting external classifier information for the LGC and WvRN methods, and using least squares (LS) divergence measure
(referred to as the LSR method) as opposed to the KL-divergence measure used in the IR methods. We also establish the relation between the LSR method and the IR methods.

Our problem may be solved in two different settings. In the first setting we select a subset of nodes for which we have high confidence in the class labels and fix the labels of these nodes. For example we may select the nodes based on thresholding its probability or decision function score above a certain value. We can treat the remaining nodes as unlabeled and solve it as a transduction problem in which relational information is used to {\it propagate} labels from the labeled set to the unlabeled set through the connections according to their strengths. In the second setting we make use of the external class distribution of all the nodes fully and no nodes are treated as unlabeled. For different methods this is done in different ways.

Detailed experiments on four benchmark datasets indicate that the second solution setting (full use of class distribution information) is better. Further experiments (in setting 2) on real world shopping domain datasets clearly demonstrate that significant performance gain can be achieved with the proposed methods. Considering speed, robustness and accuracy together, the experimental results indicate that the LSR and WvRN extension methods are better than the other methods.

The paper is organized as follows. Section 2 formulates the problem and describes the two solution settings. The methods are grouped as function estimation and probability distribution based methods; details on how they are modified for the two settings are given in sections 3 and 4. In section 5 we give experimental results, and summarize key observations in section 6.



\section{Solution Settings.}

In this section we present the statement of the problem and discuss two solution settings in which the problem can be solved. Then we briefly mention the methods and their extensions considered in this paper.

\subsection{Notations and Problem Formulation.} 

Assume that we have a graph $G$ with vertices $V$ and edges $E$. Let the vertices (nodes) $v_i,~i\in N$ where $N=\{1,\ldots,n\}$ represent the entities.
Let the edges $e_{i,j}$ (where $i,j\in N$) encode relational information between node $v_i$ and $v_j$, with weight $w_{i,j}$. Let ${\bf W}$ denote the weight matrix with $w_{i,j}$ as its elements. Let $\Delta$ be the graph Laplacian matrix defined as $\Delta={\bf D}-{\bf W}$ \cite{zhu} and let its unnormalized and normalized versions be denoted as: ${\bf L}_{un}~=~\Delta$ and ${\bf L}_{nrm}~=~{\bf D}^{-\frac{1}{2}}\Delta{\bf D}^{-\frac{1}{2}}~=~{\bf I}-{\bf D}^{-\frac{1}{2}}{\bf W}{\bf D}^{-\frac{1}{2}}$ respectively. ${\bf D}$ is a diagonal matrix with $i$th diagonal element defined as $d_{ii}~=~\sum_{j}w_{i,j}$ and $d_{ii}$ measures degree of $i$th node. 
In a traditional problem formulation, labeling of $v_i$ corresponds to specifying its class label $c_i$ where $c_i\in\{1,\ldots,K\}$ and $K$ is the number of classes. A related quantity is the vector ${\bf y}_i=\delta_{k,c_i}$ where $\delta$ is the Kronecker delta function. Let ${\bf Y}=[{\bf y}_1,\ldots,{\bf y}_n]$. Alternatively, one can specify the class distribution information ${\bf p}_i$ where ${\bf p}_i=[p_{i,1},\ldots,p_{i,K}]^T$ with $\sum_{k}p_{i,k}=1$. Let ${\bf P}=[{\bf p}_1,\ldots,{\bf p}_n]$.
The case where only class label information ($c_i$, ${\bf y}_i$ and ${\bf Y}$) are given may be viewed as a special case of the class distribuion view with ${\bf P}={\bf Y}$.
Therefore without loss of generality we assume that we are given ${\bf P}$ and we can derive the class label for node $v_i$ as $c_i={argmax \atop k} p_{i,k}$.
Thus, given ${\bf P}$, we can obtain derived labels $c_i$ and obtain the corresponding performance (e.g., accuracy) $Perf({\bf P})$. In this paper we are mainly concerned with effective use of an initial ${\bf P}$ (that is available from some external means) and using it together with the relational information to do better. In such a case we call $Perf({\bf P})$ as
the {\it initial performance} of the classifier. The classification problem of using relational information (${\bf W}$) and initial class distribution (${\bf P}$) can be loosely stated as follows: given $(G,{\bf W},{\bf P})$, find ${\hat {\bf P}}$ such that $Perf({\hat {\bf P}})$ is better than $Perf({\bf P})$. Given the above formulation we consider two settings in which the problem can be solved.

\subsection{Solution Setting 1.} In the first setting the external class distribution information available with all the nodes (${\bf P}$) is first used to select a subset of labeled nodes, ${\bf S}$. We treat the remaining nodes as unlabeled.
Then, for each $i\in {\bf S}$ we fix the label of node $i$ as $\arg\max_k p_{i,k}$.
Once this is done, the problem can be solved using any standard graph based transductive or semi-supervised learning method. However, while conventional semi-supervised learning or transduction problem settings usually assume that ${\bf p}_i$ or ${\bf y}_i$ is accurate, this is not the case here. The selection of subset of nodes is an important aspect and will be addressed later when we discuss specific methods.

\subsection{Solution Setting 2.} In the second setting the external class distribution information of {\it all} the nodes is used in the solution process. Exactly how this is done will be detailed in later sections.

\subsection{Existing Methods and Extensions.} Graph based transduction methods fall in one of two groups: (1) those which are based on function estimation and (2) those which are based on probability distribution estimation. In this paper we consider the following methods from each group: the Gaussian Field Harmonic Function (GFHF) method \cite{zhu} and the Local-Global Consistency (LGC) method \cite{zhou} from the first group and the Information Regularization (IR)~\cite{cord}, dual IR (DIR)~\cite{koji} methods and the probabilistic Weighted vote Relational Neighbor (WvRN) Classification method with Relaxation Labeling \cite{sofus} from the second group.
While all these five methods can easily solve the problem in the first setting, the second setting is (briefly) discussed in the literature only for GFHF, IR and DIR methods. In the following sections we extend LGC and WvRN methods to handle the second setting. We also propose an efficient least squares regularization (LSR) based method as compared to the KL-divergence based IR methods and relate these methods. 
\section{Function Estimation Methods}
Graph based classification methods in this group find the classification function by minimizing the function:
\begin{equation}
		Q({\bf F})~=~Q_{GR}({\bf F})~+~C Q_{datafit}({\bf F},{\bf Y})
\label{Objfn1}
\end{equation}
where $C$ is a positive regularization constant. The objective function consists of two terms.
The first term known as the {\it graph regularization} (GR) term makes use of the weight matrix ${\bf W}$ and imposes smoothness of the function over the graph.
The second term is dependent on the known label information and measures deviations from {\it model implied} label information. Consequently, this term is often called the {\it data fitting} term. In the traditional transductive setting, labels for a subset of nodes, ${\bf S}$ are given and the remaining labels need to be inferred. For example, in the GFHF method, the function ${\bf F}$ is set to ${\bf Y}$ for the labeled nodes ${\bf S}$ and is estimated for the unlabeled nodes by minimizing only the graph regularization term ${\bf Q}_{GR}({\bf F})=\sum_{k=1}^K {\bf F}^T_k{\bf L}_{un} {\bf F}_k$. In the LGC method, the function ${\bf F}$ is estimated by minimizing $C \sum_{k=1}^K {\bf F}^T_k{\bf L}_{nrm}{\bf F}_k~+~||{\bf F}_k-{\bf Y}_k||^2$ where ${\bf Y}$ is set to zero for the unlabeled nodes. 

As explained in section 2, applying the methods in solution setting 1 is straightforward. The selection of $S$ will be discussed below. Some care is needed when extending the methods to deal with solution setting 2, which aims to make effective use of the class distribution information ${\bf P}$. It makes good sense to set ${\bf Y}={\bf P}$ in (\ref{Objfn1}) and thus try and force ${\bf F}$ to be close to ${\bf P}$ while also ensuring its smoothness on the graph. It is also a good idea to try a bit more and use the uncertainty present in the class distribution information to apply different weights to different data fitting terms. With this in mind we give a generic quadratic formulation with two parameters (${\bf H}$ and $\Lambda$). Original versions of the GFHF method and the LGC method follow as special cases. Also, choosing the parameters differently using ${\bf P}$ leads to extensions of these methods to solution setting 2.


\subsection{Generic Quadratic Objective Function.} We define the following generic quadratic objective function to estimate ${\hat {\bf F}}$ and we shall see that the objective functions used in GFHF and LGC methods are special cases of this objective function.
\begin{equation}
		Q_{g}({\bf F})~=~\sum_{k=1}^K C ({\bf F}_k-{\bf V}\Lambda{\bf Y}_k)^T{\bf H}({\bf F}_k-{\bf V}\Lambda{\bf Y}_k)~+{\bf F}^T_k{\bf L}{\bf F}_k
\label{GenObjFn}
\end{equation}
where ${\bf F}=[{\bf F}_1 \cdots {\bf F}_K]$ and $Y_k$ is $k$th column of ${\bf Y}$. On comparing (\ref{GenObjFn}) and (\ref{Objfn1}) we see that the data fitting term is nothing but a weighted quadratic error function and the graph regularization term is another quadratic function defined in terms of graph Laplacian matrix ${\bf L}$. 


Let us consider the weighted quadratic error function. We have introduced two parameters: a generic error weight matrix ${\bf H}$ and a label degree matrix $\Lambda$. The role of ${\bf H}$ is to give different weightage to individual errors (data fitting terms) and is assumed to be positive definite. When ${\bf H}$ is diagonal the data fitting term is essentially a weighted sum of squared errors with error in the $i$th node weighted by $h_{ii}$. $\Lambda$ is a diagonal matrix and its role is to incorporate any label degree information we want to associate with each node where $0\le \lambda_{ii}\le 1$. For example if node $i$ is unlabeled then $\lambda_{ii}=0$ and when it is labeled {\it fully} $\lambda_{ii}=1$. Note that the interval $0<\lambda_{ii}<1$ brings in the notion of {\it partial} labeling and provides flexibility when ${\bf y}_i$ ($i$th row of ${\bf Y}$) is inaccurate.


Finally, ${\bf V}$ is a node regularization matrix which is also diagonal. Wang et al\cite{wang} introduced the notion of node regularization in the original LGC formulation to handle {\it label imbalance} problem in the graph. Assuming ${\bf y}_i$ to contain only one non-zero element they defined normalized label matrix ${\tilde {\bf Z}}={\tilde V}{\bf Y}$ with $v_{ii}=\sum_{k=1}^K \frac{y_{i,k}d_{ii}}{\tilde \eta_k}$. Here ${\tilde \eta}_k=\sum_{i=1}^n d_{ii}y_{i,k}$ and we can see that $\sum_{i=1}^n {\tilde z}_{i,k}=1$. Thus ${\bf V}$ balances the influence of labels from different classes and allows with high degree to make more contribution. In our general setting we allow $0\le y_{i,k}\le 1$, $k=1,\ldots,K$ subject to $\sum_{i=1}^K y_{i,k}=1$ and we also have the label degree matrix $\Lambda$. Therefore we redefine $v_{ii}$ as: $v_{ii}=\sum_{k=1}^K \frac{d_{ii} \lambda_{ii} y_{i,k}}{\eta_k}$ where $\eta_k=\sum_{i=1}^n d_{ii}\lambda_{ii}y_{i,k}$. Defining ${\bf Z}={\bf V}\Lambda{\bf Y}$ we have $\sum_{i=1}^n z_{i,k}=1,~\forall k$ and, ${\bf Z}$ becomes a normalized matrix as earlier.

Given the additive nature of the function ${\bf Q}_{g}({\bf F})$, ${\bf F}_k$ can be solved independently. Then setting the derivative ${{\partial {\bf Q}_g({\bf F})} \over {\partial {\bf F}_k}}$ to zero gives the following closed form solution that involves matrix inversion and costs $O(n^3)$.
\begin{equation}
	{\hat {\bf F}}_k=C\bigl({\bf L}+C{\bf H}\bigr)^{-1}{\bf H}{\bf Z}_k
\label{CFSol}
\end{equation}
Here ${\bf Z}_k$ is $k$th column of ${\bf Z}$. In the following discussion we assume that ${\bf H}$ is diagonal and is invertible. Then equation (\ref{CFSol}) can be rewritten as: ${\hat {\bf F}}_k=\bigl({\bf I}+\frac{1}{c}{\bf H}^{-1}{\bf L}\bigr)^{-1}{\bf Z}_k$. Then instead of estimating ${\hat {\bf F}}_k$ from (\ref{CFSol}) using matrix inversion, we can find it by iteratively using the fixed point equation: ${\hat {\bf F}}=-\frac{1}{c}{\bf H}^{-1}{\bf L}{\hat {\bf F}}+{\bf Z}$. Note that the convergence rate of the iterative solution depends on the eigen values of the matrix $\bigl({\bf I}+\frac{1}{c}{\bf H}^{-1}{\bf L}\bigr)$. Also the fixed point equation changes depending on whether normalized or unnormalized graph Laplacian is used and the specific form of error weight matrix ${\bf H}$.


\subsection{GFHF Method As a Special Case.} The GFHF method \cite{zhu} was originally proposed to solve semi-supervised learning problem based on a Gaussian random field model for a weighted graph model. In this formulation a real valued function ${\bf f}$ is estimated by minimizing a quadratic energy function ($\sum_{k=1}^K {\bf F}^T_k{\bf L}_{un} {\bf F}_k$) subject to the constraint that the function ${\bf f}$ takes the actual label values on the set of labeled nodes. It is assumed that $y_{i,k}$ takes value in $\{0,1\}$. This solution is retrieved by considering our generic quadratic formulation and setting ${\bf L}={\bf L}_{un}$, ${\bf V}={\bf I}$, ${\bf H}={\bf D}\Lambda({\bf I}-\Lambda)^{-1}$ and $C=1$. Further $\lambda_{ii}=1$ if node $i$ is labeled and is zero otherwise.
Note that the specific structure of ${\bf H}$ and $\Lambda$ imposes the constraints on the labeled nodes by assigning infinite weights to the errors on labeled nodes. On substituting these specific matrices we get:
\begin{equation}
{\bf F}_k=\bigl({\bf I}-\bigl({\bf I}-\Lambda\bigr){\bf D}^{-1}{\bf W}\bigr)^{-1}\Lambda{\bf Y}_k
\label{GFHFEqn}
\end{equation}
where ${\bf Y}_k$ is $k$th column of the matrix ${\bf Y}$. The corresponding fixed point equation (after including the node regularization matrix ${\bf V}$) is: ${\bf F}_k=\bigl({\bf I}-\Lambda){\bf D}^{-1}{\bf W}{\bf F}_k+\Lambda{\bf Z}_k$.

Note that (\ref{GFHFEqn}) can handle both the solution settings presented in section 2. In a transductive setting with labeled and unlabeled nodes we have $\lambda_{ii}=1$ for the {\it labeled} subset of nodes such that $i \in {\bf S}$ and $\lambda_{ii}=0$ for the remaining set of nodes $i \in {\bar {\bf S}}$. \cite{zhu} also gave a random walk interpretation to this method. That is, ${\hat f}_{i,j}$ has an interpretation as the probability that a particle starting from node $i$, hits a labeled node with label $1$ (with class $j$ being considered as $1$ like in one-vs-all setting). They also discussed incorporating external class label information using the notion of dongle nodes. Here each node having label information is attached with a dongle node and the label value is tied to this node. Further a transition of probability $\mu$ was defined from each node to its dongle node and all other transitions are covered with probability $(1-\mu)$.

It may be noted that (\ref{GFHFEqn}) is a general form with this interpretation and when $\Lambda=\mu{\bf I}$ we get the original form of GFHF method. Thus we allow these transition probabilities to take different values for different nodes and this is very useful since we may have different levels of confidence on the label information we get for each node from the external classifier. Below we shall discuss how to choose these values.

\subsection{LGC Method As a Special Case and its Extension.} Zhou et al\cite{zhou} proposed a semi-supervised learning approach to design a {\it smooth} classification function ${\bf F}$ such that it respects any intrinsic structure revealed by the labeled and unlabeled nodes. Starting with the intution of label propagation they introduced an iterative algorithm which essentially finds the solution to a FP equation. With ${\bf y}_i$ having only one non-zero element and by setting, in our generic quadratic formulation, ${\bf L}={\bf L}_{nrm}$, ${\bf V}={\bf I}$, ${\bf H}={\bf I}$, $\lambda_{ii}=1,~\forall i \in {\bf S}$ and $\lambda_{ii}=0,~\forall i \in {\bar {\bf S}}$ we can retrieve the FP equation following the same steps in the previous section. Zhou et al\cite{zhou} also considered several variants of fixed point equations and showed how one such FP equation can be obtained as solution from optimizing a regularized objective function ($C \sum_{k=1}^K {\bf F}^T_k{\bf L}_{nrm}{\bf F}_k~+~||{\bf F}_k-{\bf Y}_k||^2$).  Equation (\ref{GenObjFn}) can be seen as a generalized version of such an objective function. Unlike GFHF method the regularization constant $C$ is to be set using standard techniques like cross-validation and this is possible only when sufficient number of labeled nodes is available.

{\bf Extension for Setting 2} We extend the basic LGC method to handle setting 2 as follows. Firstly, we relax ${\bf y}_{i}$ as described earlier. Secondly, we make use of the label degree matrix $\Lambda$ to handle inaccurate label information. Finally, we allow ${\bf S}$ to be the entire set of nodes. With specific structures of ${\bf L}$ and ${\bf H}$ defined as above we get:
\begin{equation}
{\bf F}_k=(1-\gamma)\bigl({\bf I}-\gamma{\bf D}^{-\frac{1}{2}}{\bf W}{\bf D}^{-\frac{1}{2}}\bigr)^{-1}\Lambda{\bf Z}_k
\label{LGCEqn}
\end{equation}
where $\gamma=\frac{1}{1+C}$ and it can re-written in fixed point form as done earlier with the GFHF method. We will discuss shortly how to choose $C$ and $\Lambda$ in the two settings.

The solution of the fixed point equations, (\ref{GFHFEqn}) and (\ref{LGCEqn}) can be obtained iteratively. In all our experiments we found the solution iteratively. The iterative update is very useful when ${\bf W}$ is sparse (which is often the case in practice) and the convergence rate is dependent on the eigen values of the matrix involved in the inversion. In the worst case the cost is expected to be in the order of $O(mn^2)$ where $m$ is a bound on the number of neighbors of a node in the graph. When the graph is dense, the complexity can be cubic in $n$ \cite{bengio}. 



\subsection{Parameter Selection in Two Settings.} In this section we present ways of selecting the subset ${\bf S}$ and other parameters $\Lambda$ and $C$ that appear in (\ref{GenObjFn}).
We propose two scoring schemes that can be used to make the subset selection for solution setting 1.

{\bf Maximum Probability Scoring Scheme} Let $p^{max}_i=\max_l~p_{i,l}\:\forall i$. Then we sort the nodes based on the maximum probability score ($p^{max}_i$) of each node from its class distribution. Then the subset of nodes that satisfy the condition $p^{max}_i\ge p_{th}$ can be selected as the set of labeled nodes. Essentially we select nodes on which we have high confidence in their labeling assignments. If $p_{th}$ is very high then the number of labeled nodes becomes smaller. However the noise level (percentage of noisy labeled nodes) will also be low and care has to be taken to ensure that each class has at least some labeled nodes. On the other hand, if the threshold level is low then though the number of labeled nodes increases the noise level also increases. Alternately, we can also select top-M percentage of nodes as the subset of labeled nodes. We will refer to the subset selection scheme based on the maximum probability score as $MPS$ scheme.

{\bf Entropy based Scoring Scheme} In this scheme we sort the nodes based on an entropy based score $\eta_i$ for node $i$ defined as:
	$\eta_i~=~1 - E({\bf p}_i)/E_{max}$  
where $E({\bf p}_i)$ represents the entropy of the class distribution ${\bf p}_i$ and $E_{max}=\log(K)$ is the maximum entropy possible for a given number of classes $K$. Thus $\eta_i$ lies in the interval $[0,1]$ and takes high or low values depending on the spread of the class distribution. This score is motivated from the view that we do not have any class label information when the class distribution is uniform and we are certain when ${\bf p}_{i,k}=\delta_{k,c_i}$. Setting a threshold to select the subset of labeled nodes ${\bf S}$ is harder here compared to $MPS$ scheme and as mentioned earlier, so we can simply select the top-M percentage of nodes sorted by this score. We will refer to this subset selection scheme as $EBS$ scheme.

Note that the above two schemes only use the class distribution information to select the subset of nodes. It is also possible in addition to make use of the graph structure information and is a direction for further work.

{\bf Choice of $\Lambda$} Recall that $\lambda_{ii}$ has the interpretation of transition probability in GFHF method; and it has the interpretation of {\it partial} label information in the LGC-extension method. Earlier we gave two useful scoring schemes based {\it only} on the external classifier information. We can make use of either of these schemes to fix $\lambda_{ii}$. That is, we can either set $\lambda_{ii}=p^{max}_i$ or $\lambda_{ii}=\eta_i$. We evaluate both these schemes in our experiments. Note that when the subset selection scoring scheme also makes use of graph structure information, then we may not want to use the same scoring scheme to choose $\lambda$. This is because we may like to give importance to the label degree or transition probability to each node {\it only} based on the information from the external classifier.


{\bf Choice of C} We use 5-fold cross-validation (CV) technique to choose $C$ for LGC. In the first setting, for each value of $C$ chosen from a range in the log scale, we evaluate 5-fold CV accuracy on the selected subset of nodes using the label information available for these nodes. Then we pick the value of $C$ that gives the maximum accuracy and find the final solution using the chosen $C$ value. Here the label information is nothing but the label that has the maximum probability score. In the second setting we evaluate 5-fold CV accuracy on top-M percentage of nodes selected using MPS or EBS scheme.
It is worth pointing out that, inaccuracies in the external class distribution could lead to a choice of $C$ that is tilted away from the best possible choice.

\section{Probabilistic Methods}

The methods in this class, viz., the information regularization (IR), dual IR methods and the probabilistic weighted vote relational neighbor (WvRN) classifier method, estimate the class distributions of the nodes. In this section we discuss their adaptation for the two solution settings and also introduce least squares regularization (LSR) based method. Overall, we group the IR, DIR and LSR methods under the category of region based regularization methods. 


\subsection{WvRN Method and its Extension.} The original probabilistic weighted vote relational classifier (with relaxation labeling) method \cite{sofus} was formulated to solve the collective classification problem where class distributions of a subset of nodes are known and fixed. Then the class distributions of the remaining (unlabeled) nodes are obtained by an iterative algorithm. We give a version of this algorithm in Algorithm \ref{alg:WvRN}. It has two components, namely, weighted vote relational neighbor classifier (WvRN) component and relaxation labeling (RL) component. The relaxation labeling component performs collective inferencing and keeps track of the current probability estimates ${\bf p}^{(t)}_i$ for all unlabeled nodes at each time instant $t$. These {\it frozen} estimates ${\bf p}^{(t)}_i$ are used by the relational classifier. The relational classifier computes the probability distribution for each unlabeled node as the weighted sum of probability distributions ${\bf p}^{(t)}_j$ of its neighbors with weight $w_{ij}$. Since relaxation labeling may not converge, simulated annealing is performed to ensure convergence (as given in step (2) of the algorithm)~\cite{sofus}. Note that $\beta^{(t)} \rightarrow 0$ for sufficiently large number of iterations; therefore, convergence is guaranteed. It has been observed that the performance is robust when $0.9<\nu<1$. The WvRN algorithm (Algorithm \ref{alg:WvRN}) can be directly used to solve the problem in the first setting by setting the probability distributions associated with the selected set of nodes $i \in {\bf S}$ to $\delta_{k,c_i}$ and initializing the probability distributions of the remaining set of nodes (${\bar {\bf S}}$) with the class prior obtained using the label information of the selected set of nodes. Here the label for a node ($c_i$) in the selected subset is defined as the label with the maximum probability score obtained from $p_i$. 

{\bf Extensions for Setting 2} Next we extend the WvRN algorithm to solve the problem in setting 2. We consider two variants. In the first variant we consider a simpler form where we initialize ${\bf p}^{(0)}_i$ with the external classifier information for all nodes and run Algorithm \ref{alg:WvRN} as it is. In the second variant we use the dongle node idea (used in the GFHF method) and modify the relational classifier estimate from ${\bf q}_i$ to ${\tilde {\bf q}}_i$ as follows. With $\lambda_{ii}$ representing the transition probability, we define ${\tilde q}_{i,k}=\lambda_{ii} p^{(0)}_{i,k}+(1-\lambda_{ii})q_{i,k}$ (where $q_{i,k}$ is as defined in Algorithm \ref{alg:WvRN}) and ${\bf p}^{(0)}_i$ is the distribution information available from the external classifier. We can select $\Lambda$ using MPS or EBS schemes, and $\nu$ using the cross-validation technique (see section 3.4).


\begin{Algorithm}
	\caption{Probabilistic WvRN Algorithm}
	\label{alg:WvRN}
\begin{compactitem}
	\item Set $t=0$, $\beta^{(0)}=1$ and $\nu$=0.95
	\item For all nodes $i \in {\bar {\bf S}}$, initialize ${\bf p}^{(0)}_i$ to class prior (obtained from known labeled nodes)
	\item Until convergence holds for all the nodes in ${\bar {\bf S}}$ do:
        \begin{description}
	      \item For each element $i \in {\bar {\bf S}}$ and $k=\{1,\ldots,K\}$
	      \begin{enumerate}
              \item Estimate node class probability (using neighbor information) $q_{i,k}=\frac{1}{\psi}\sum_{j} w_{i,j} p^{(t)}_{j,k}$ (where $\psi$ is a normalizing constant)
			\item Set $p^{(t+1)}_{i,k}=\beta^{(t)}q_{i,k}+(1-\beta^{(t)})p^{(t)}_{i,k}$
              \end{enumerate}
              \item Set $t=t+1$ and $\beta^{(t+1)}=\beta^{(t)}*\nu$
		\end{description}
\end{compactitem}
\end{Algorithm}

\subsection{Region Based Regularization Methods}
Corduneanu and Jakkola~\cite{cord} proposed the information regularization method. They introduced a notion of region, where each region is defined as a subset of nodes in the graph. The intuition is that nodes belonging to a given region have the same label. Here a weight is associated with each region and a weight to each node that defines the relative importance of points that belong to a given region. Further, a distribution is associated with each region and node. Then the distributions of labels are propagated on a graph for semi-supervised or transductive learning. We refer to methods that work within this framework of region as region based regularization methods. The IR, DIR and LSR methods fall in this category. In graph based classification problem, each edge forms a region and the weight of the region is nothing but the edge weight; further, an equal degree of importance is given to both the nodes connected by the edge. Then the regularized optimization problem can be written as minimization of the objective function: 

\begin{equation}
		H({\bf P};{\bf W}) = C\sum_{i \in {\bf S}} \lambda_{ii} D({\bf p}^{(0)}_i, {\bf p}_i) +  \sum_{i,j} w_{i,j} D({\bf p}_j,{\bar {\bf p}}_{i,j})
\label{RegObjFn1}
\end{equation}
where $D(\cdot)$ denotes a divergence measure, $\lambda_{ii}$ denotes {\it relative} weight factor that we would like to assign to $i$-th node in ${\bf S}$. 
Further, ${\bar {\bf p}}_{i,j}$ denote the probability distribution associated with the region (edge) $(i,j)$. The optimization problem should include the following constraints: (1) $0\le p_{i,k}\le 1$ and (2) $\sum_{k=1}^K p_{i,k}=1$. The region distributions are constrained in a similar way. In this formulation, an alternating learning algorithm\cite{cord} consists of two steps. In the first step, {\it all} the node distributions ${\bf p}_i, \forall i$ collected as a vector ${\bf P}$ are fixed and, {\it all} the region distributions ${\bar {\bf p}}_{i,j},\forall (i,j)$ pairs are obtained by minimizing {\it only} the second term in (\ref{RegObjFn1}). In the second step, using these estimated region distributions, the node distributions are re-estimated by minimizing {\it both} the terms in (\ref{RegObjFn1}). These steps are repeated until convergence.

Now let us look at how (\ref{RegObjFn1}) is used in the two solution settings. In the first setting, having selected the subset of nodes ${\bf S}$ we fix $p_{i,k}=p^{(0)}_{i,k}=\delta_{k,c_i},~\forall i \in {\bf S}$ and $k=1,\ldots,K$. Then the optimization is over ${\bf P}_{{\bf V} \setminus {\bf S}}$ where we have explicitly indicated the set of nodes to be optimized. Therefore the first term in (\ref{RegObjFn1}) does not play a role in the optimization process. In the second setting, the first term also plays a role since 
we optimize over the entire set of node distributions ${\bf P}_{\bf V}$. Using this broad setup let us consider several methods that stem from different divergent measures $D(\cdot)$, namely, KL-divergence and squared error (loss). The complexity of all these methods is same as that of the function estimation methods.      
        
\subsubsection{Information Regularization (IR) Method}
Within the above framework, Corduneanu and Jakkola~\cite{cord} used KL-divergence as the divergence measure $D(\cdot)$ and called the second term as {\it information regularization} term. Then closed form solution can be obtained for ${\bar {\bf p}}_{i,j}$ and is given by: 
\begin{equation}
{\bar {\bf p}}_{i,j} =\frac{1}{2}({\bf p}_i+{\bf p}_j).
\label{regdist1}
\end{equation}
Note that ${\bar {\bf p}_{i,j}}$s are the region (edge) distributions and are obtained from minimizing the second term in (\ref{RegObjFn1}) with the KL-divergence measure. 
As mentioned earlier, in solution setting 1 the first term does not play a role in the optimization process. Then the distributions for the unlabeled nodes are directly obtained as: 
\begin{equation}
p_{i,k}=\frac{1}{\chi_i}exp(\sum_{i,j}w_{i,j} \log {\bar p}_{i,j}(k))
\label{nodexpdist}
\end{equation}
where ${\bar p}_{i,j}(k)$ denotes $k$th component in ${\bar {\bf p}}_{i,j}$ (see (\ref{regdist1})) and $\chi_i$ denotes a normalizing constant. Thus in setting 1, closed form solutions exist in {\it both} the steps of the alternating learning algorithm. In solution setting 2, with the first term also playing a role in the optimization process there is {\it no closed form solution} like (\ref{nodexpdist}). Then constrained optimization using either Newton's method or exponentiated gradient algorithm is carried out. This step could be expensive and affects the scalability of the method. To address this issue, Tsuda~\cite{koji} proposed a dual information regularization method, where closed form solutions are obtained in both the steps {\it even} when the first term in (\ref{RegObjFn1}) plays a role. 

\subsubsection{Dual Information Regularization (DIR) Method} Tsuda~\cite{koji} modified the regularizer by interchanging the arguments in the KL divergence measure ($KL({\bar {\bf p}}_{i,j} || {\bf p}_i)$) and using modified ${\bar {\bf p}}_{i,j}$ given by: 
\begin{equation}
{\bar {\bf p}}_{i,j}=\frac{1}{Z_{ij}} \exp \bigl(\frac{1}{2}(\log({\bf p}_i)+\log({\bf p}_j))\bigr)
\label{regdist2}
\end{equation}
where $Z_{ij}$ is the normalization factor. Then the closed form solutions for ${\bf p}_i$ are obtained as:
\begin{equation}
{\bf p}_i = \frac{1}{\lambda_{ii}+D_{ii}}(\lambda_{ii}{\bf p}^{(0)}_i+ \sum_{i,j} w_{i,j} {\bar {\bf p}}_{i,j})
\label{nodedist2}
\end{equation}   
where $D_{ii}=\sum_{j} w_{i,j}$. We refer to the update equations (\ref{regdist2}) and (\ref{nodedist2}) as the dual information regularization ({\it DIR}) method. Let us consider the implications in the two settings. In the first setting we fix the node distributions of the nodes in ${\bf S}$ and estimate only the node distributions of the nodes ${\bf V} \setminus {\bf S}$. This estimation is done using (\ref{regdist2}) and with $\lambda_{ii} = 0, \forall i \in {\bf S}$ in (\ref{nodedist2}). In the second setting, (\ref{regdist2}) and (\ref{nodedist2}) are used as they are. Thus, closed form solutions are available in both settings and in both steps of the learning algorithm and, this helps in improving the speed of the IR method.     
  
\subsubsection{Least Squares Regularization (LSR) Method}
We propose to use the squared error as the divergence measure $D(\cdot)$; that is, for any two distributions ${\bf p}$ and ${\bf q}$, we define $D({\bf p},{\bf q})=||{\bf p}-{\bf q}||^2$. 
Then it is easy to verify from (\ref{RegObjFn1}) that the optimal ${\bar {\bf p}}_{i,j}=\frac{1}{2}({\bf p}_i+{\bf p}_j)$ and is {\it same} as the one obtained in the IR method. 
In this method, we proceed as in the 2-step algorithm where we estimate ${\bf P}_{{\bf V}}$ keeping all the node distributions fixed. This results in the closed form solution: ${\bf p}_i = \frac{1}{\lambda_{ii}+D_{ii}}(\lambda_{ii} {\bf p}^{(0)}_i+ \sum_{i,j} w_{i,j} {\bar {\bf p}}_{i,j})$. This solution is {\it exactly same} as the one given in (\ref{nodedist2}). 
We refer the updates (\ref{regdist1}) and (\ref{nodedist2}) as the least squares regularization ({\it LSR}) method. Note that application of this method in two settings is exactly same as described above for the DIR method.   

{\bf Relation to other Methods} The LSR objective function, that is, (\ref{RegObjFn1}) with squared error as the divergence measure has structure similar to the generalized quadratic function (\ref{GenObjFn}). The key difference is that unlike the rows of ${\bf F}$, the node distributions ${\bf p}_i$ and ${\bf p}^{(0)}_i$ are constrained to be probability distributions. Now, comparing the DIR and LSR updates, we see that they differ in the way region distributions are updated. {\it Thus, the LSR method interestingly combines the IR based region distribution update with the {\it D-IR} based node distribution update. Also, like the IR methods, the objective function is convex and has global minimum.} 

\begin{table}
\begin{center}
\caption{\small{Public Datasets Description. $n$, $K$, $|E|$ and $d$ denote the number of nodes, classes, edges and dimensions respectively.}}
\vskip 0.05in
\begin{small}
\begin{tabular}{|l|c|c|c|c|} \hline
 Dataset & $n$ & $K$ & $|E|$ & $d$ \\ \hline
 WebKB & 1051 & 2 & 269046  & 4840 \\ \hline
 USPS & 3874 & 4 & 6398 & 256\\ \hline
 CoraCite & 4270 & 7 & 22516 & -\\ \hline
 CoraAll & 4270 & 7 & 71824 & -\\ \hline
\end{tabular}
\end{small}
\end{center}
\label{Datasets}
\end{table}

\section{Experiments}
We conducted two experiments on the two function estimation methods, LGC and GFHF, and the four probabilistic methods, WvRN, IR, DIR and LSR. In different settings the acronyms of methods will appropriately refer to the modifications that we described earlier in the paper. For WvRN in solution setting 2 there are two versions, which we will refer to as WvRN-V1 and WvRN-V2 (see section 4.1 and 4.2).
In the first experiment we evaluate the performance of the methods in
solution settings 1 and 2 with four publicly available benchmark datasets. Next, we present results from the second experiment where we evaluated the performance in solution setting 2 on datasets constructed from web pages of three shopping sites.

\subsection{Datasets Description.} The details of the datasets are given in Table 1. The WebKB dataset contains two document categories, {\it course} and {\it noncourse}. Each document has two types of information, the webpage text content and link information. The
number of features in page and link representations are 3000 and 1840 respectively. Following \cite{wang} we constructed the graph based on cosine distance neighbors with Gaussian weights and chose 200 nearest neighbors.
The USPS dataset is a handwritten digit recognition task, for which we used the same setting as given in \cite{zhou}.
The number of features is 256, obtained from a 16 $\times$ 16 image. The four classes correspond to digits 1 to 4.
The graph is constructed using a radial basis function kernel with width set to 1.25; the number of
neighbors is set to 1. The CORA dataset comprises computer science research papers. There are seven
classes associated with the papers; the classes correspond to the following machine learning sub-topics: {\it Case-based Methods}, {\it Genetic Algorithms}, {\it Neural Networks}, {\it Probabilistic Methods}, {\it Reinforcement Learning}, {\it Rule Learning} and {\it Theory}.
The dataset consists of
the full citation graph with labels for the topic of each paper. There are two variants of this dataset,
referred to as CORACITE and CORAALL. These variants come from the way the papers are linked. 
The CoraCite variant uses {\it only} citation link and an edge
is placed between two papers if one cites the other. The weight of an edge is normally one unless the two papers cite each other, in which case it is two. The CoraAll variant uses {\it both} citation and author link information, where an edge is placed (in addition to co-citation) when two papers have author relation as well. We used the CORA dataset versions available from Netkit package described in \cite{sofus}.

\subsection{Classifier Information Generation.} While these datasets have only label information we need inaccurate external classifier (distribution) information in our problem formulation. Therefore we need a model which generates this distribution given the label information. We now describe the model used in our experiments. We fix two probability parameters $p_{min}$ and $p_{max}$ with $p_{min}<p_{max}$ that take values in $[0,1]$. Given $p_{min}$ and $p_{max}$ we generate the distribution for each node as follows. In the first step generate a random number $p_{label}$ from the interval $[p_{min},p_{max}]$ and treat $p_{c_i}$ as the probability score of the true label ($c_i)$.
Then we generate $K-1$ random numbers ($pr_k$ with $k\ne c_i)$ from the interval $[0,1]$ and assign $p_{k}=\frac{pr_k (1-p_{c_i})}{\psi}$ where $\psi$ is a normalizing constant such that $\sum_{k\ne c_i}p_k=1-p_{c_i}$. The choices of $p_{min}$ and $p_{max}$ determine the degree of inaccuracy present in the information. If $p_{min}$ is set too low many nodes get labeled wrongly. Also note that the number of classes play a role in determining the degree of accuracy since the mass $1-p_{c_i}$ gets distributed across $K-1$ number of classes. We considered different levels of inaccuracy by setting different values for $p_{min}$ and $p_{max}$. As the observations were almost the same across the methods and settings for different levels of inaccuracy, we present results only for one set of values.

\begin{figure*}[tb]
\centering
\vspace*{-21ex}
		\includegraphics[width=0.4\linewidth]{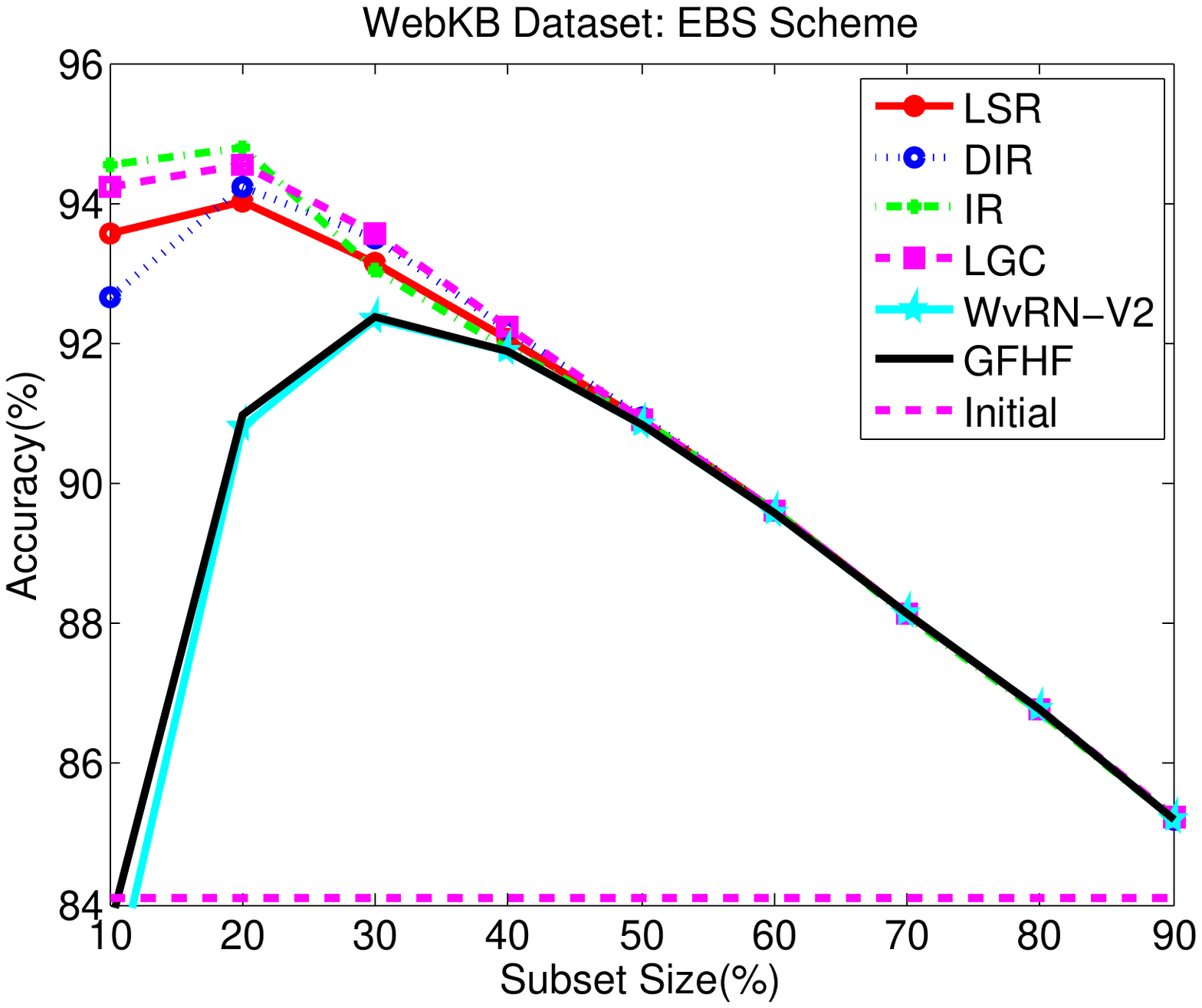}
\vspace*{-29ex}
		\includegraphics[width=0.4\linewidth]{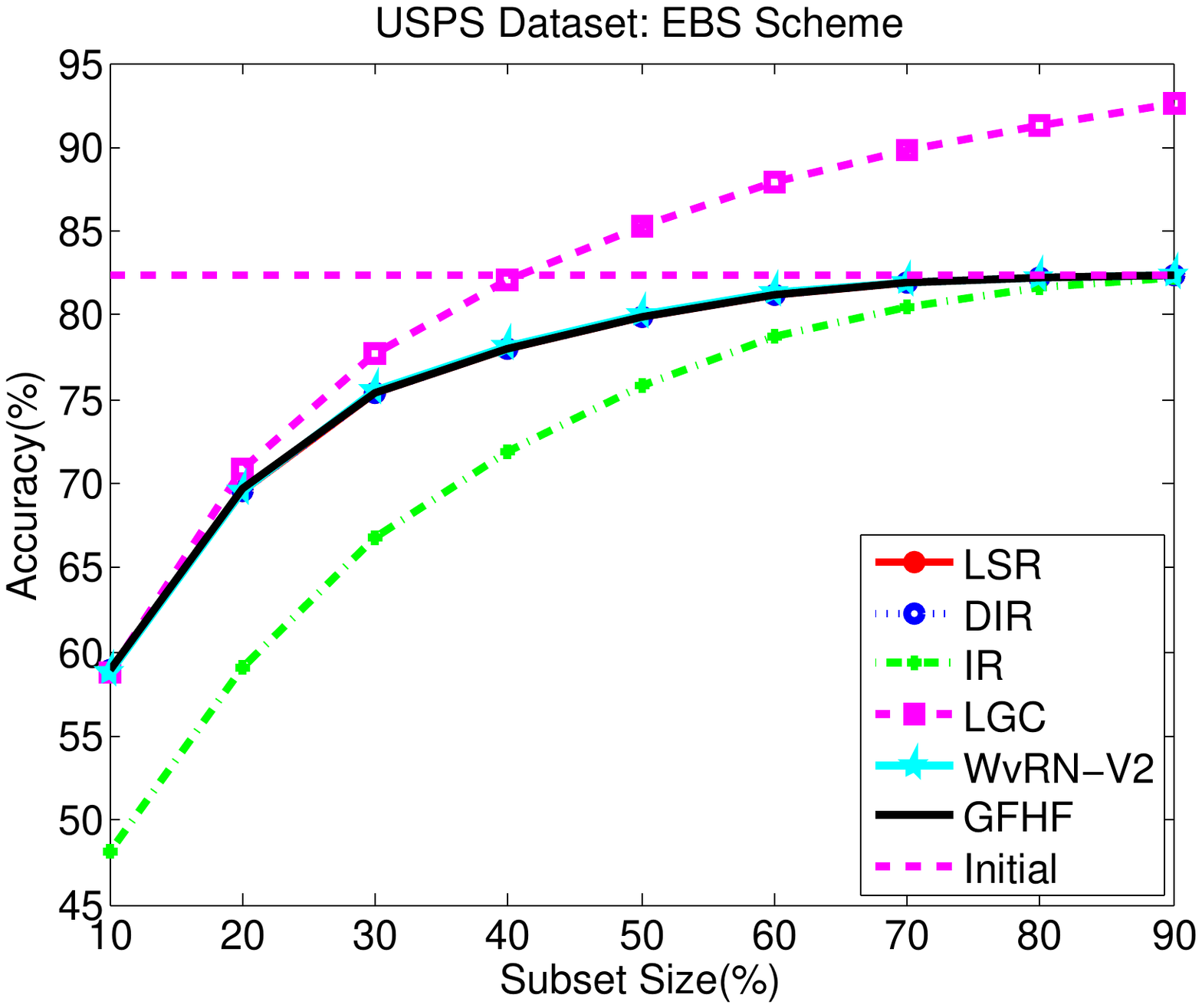} \\
		\includegraphics[width=0.4\linewidth]{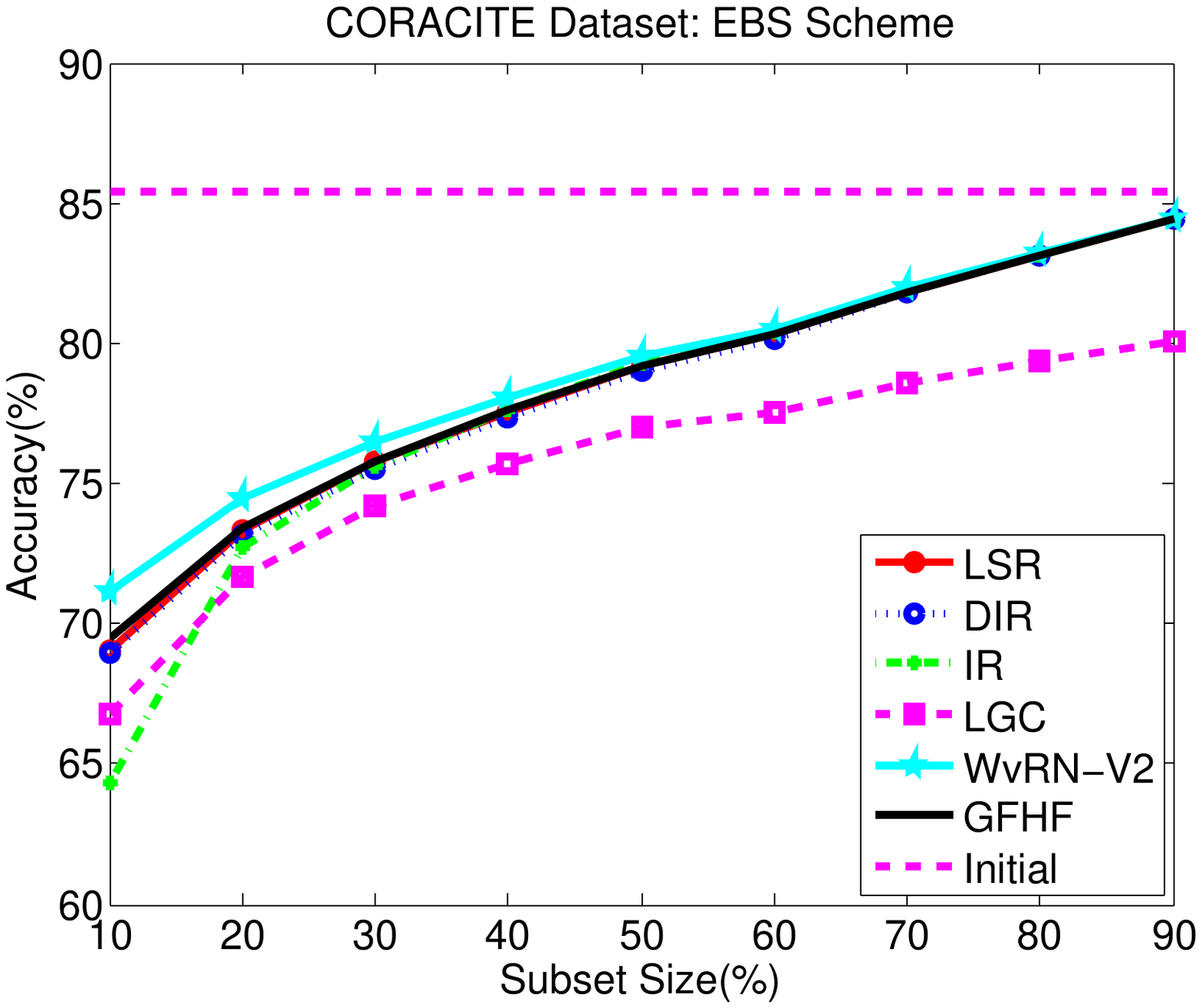}
		\includegraphics[width=0.4\linewidth]{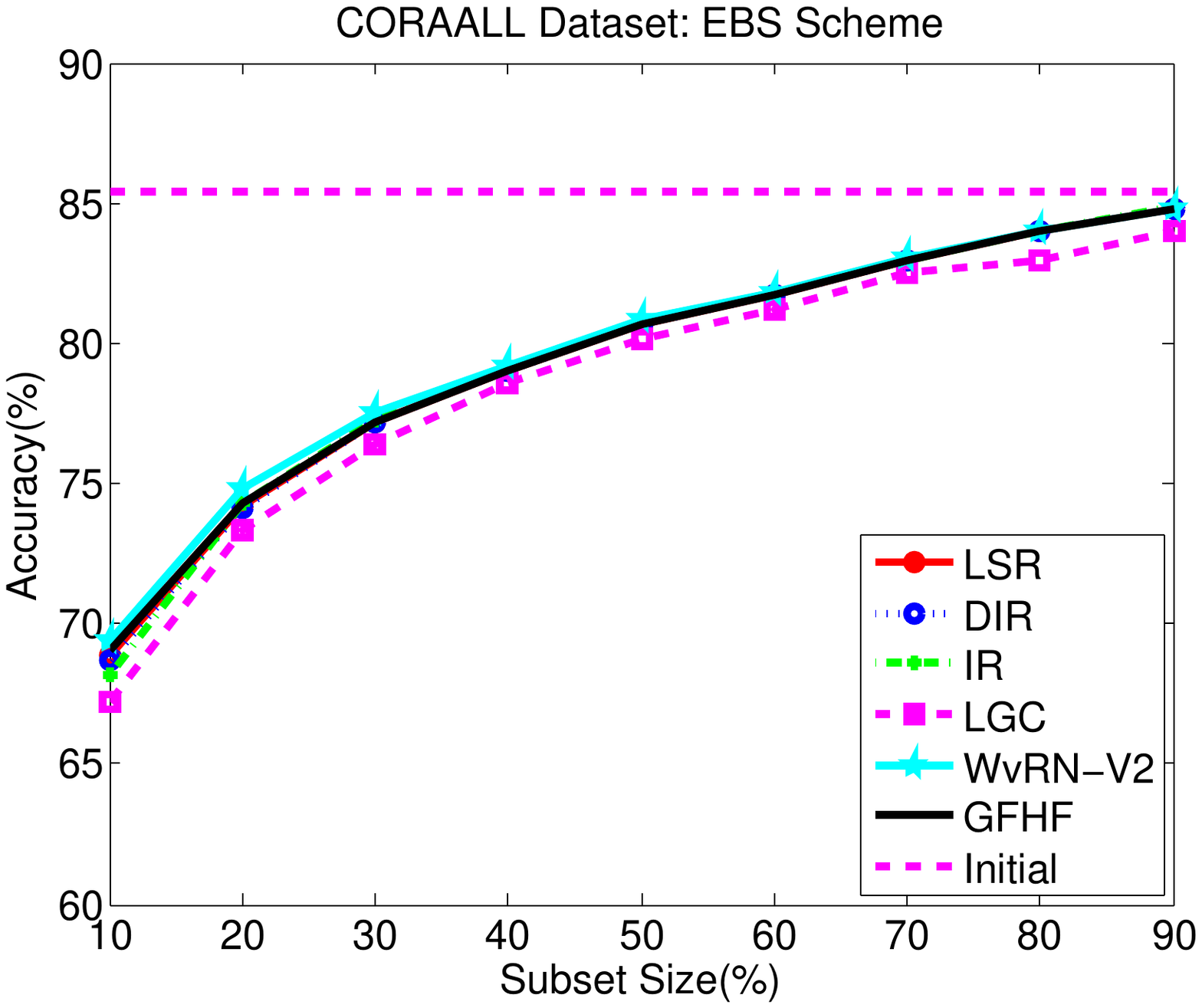}\\
\vspace*{-17ex}
 \caption{{Setting 1 Results: X-axis represents the subset size in terms of percentage of number of nodes and Y-axis represents accuracy. The results correspond to the EBS subset selection scheme. Results for the MPS subset selection scheme are not given because of their closeness with EBS; see text.  {\it Initial} refers to accuracy computed from generated inaccurate distribution.}}
\label{fig:s1}
\end{figure*}
\begin{figure*}[tb]
\centering
\vspace*{-20ex}
		\includegraphics[width=0.4\linewidth]{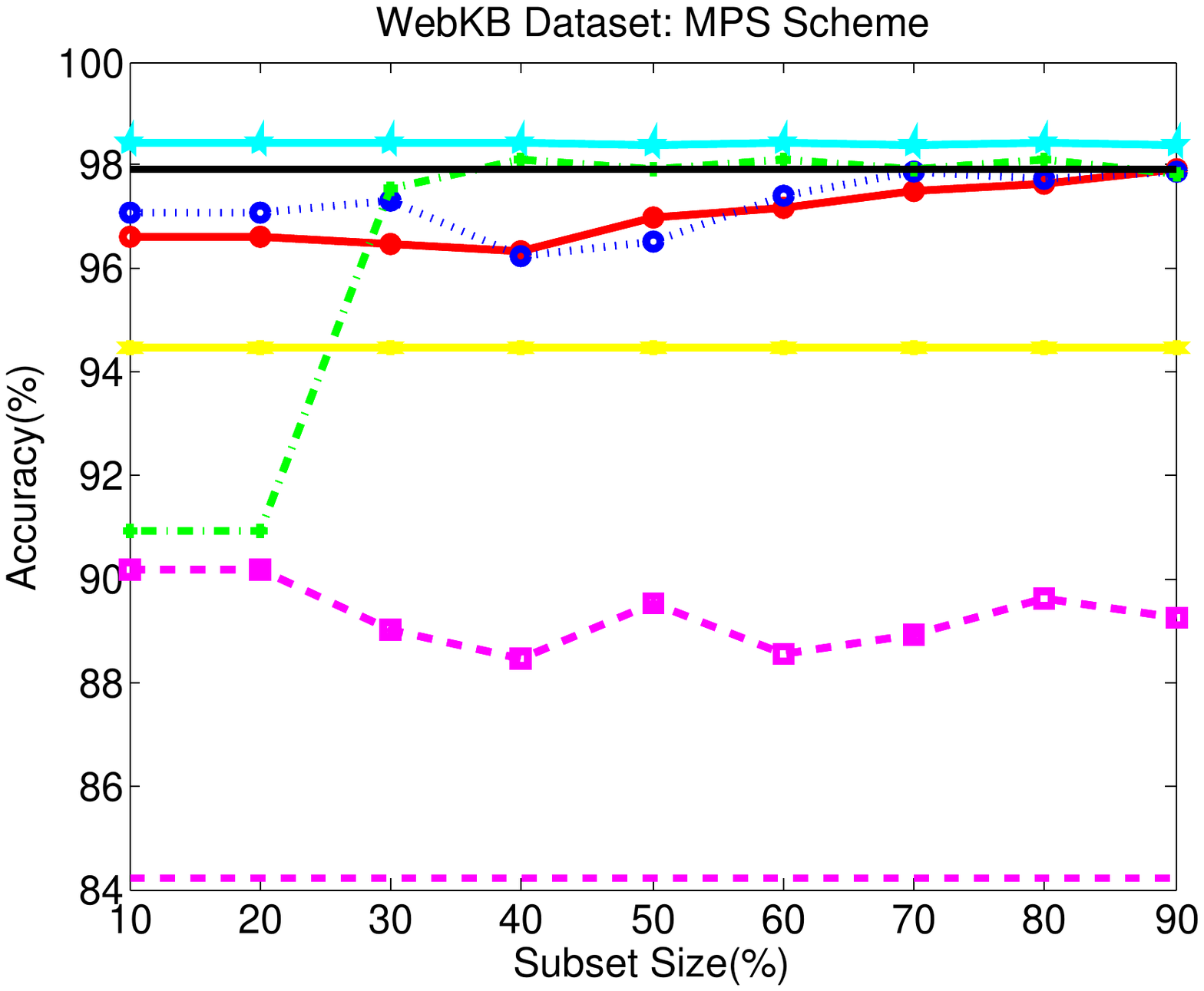}
		\includegraphics[width=0.4\linewidth]{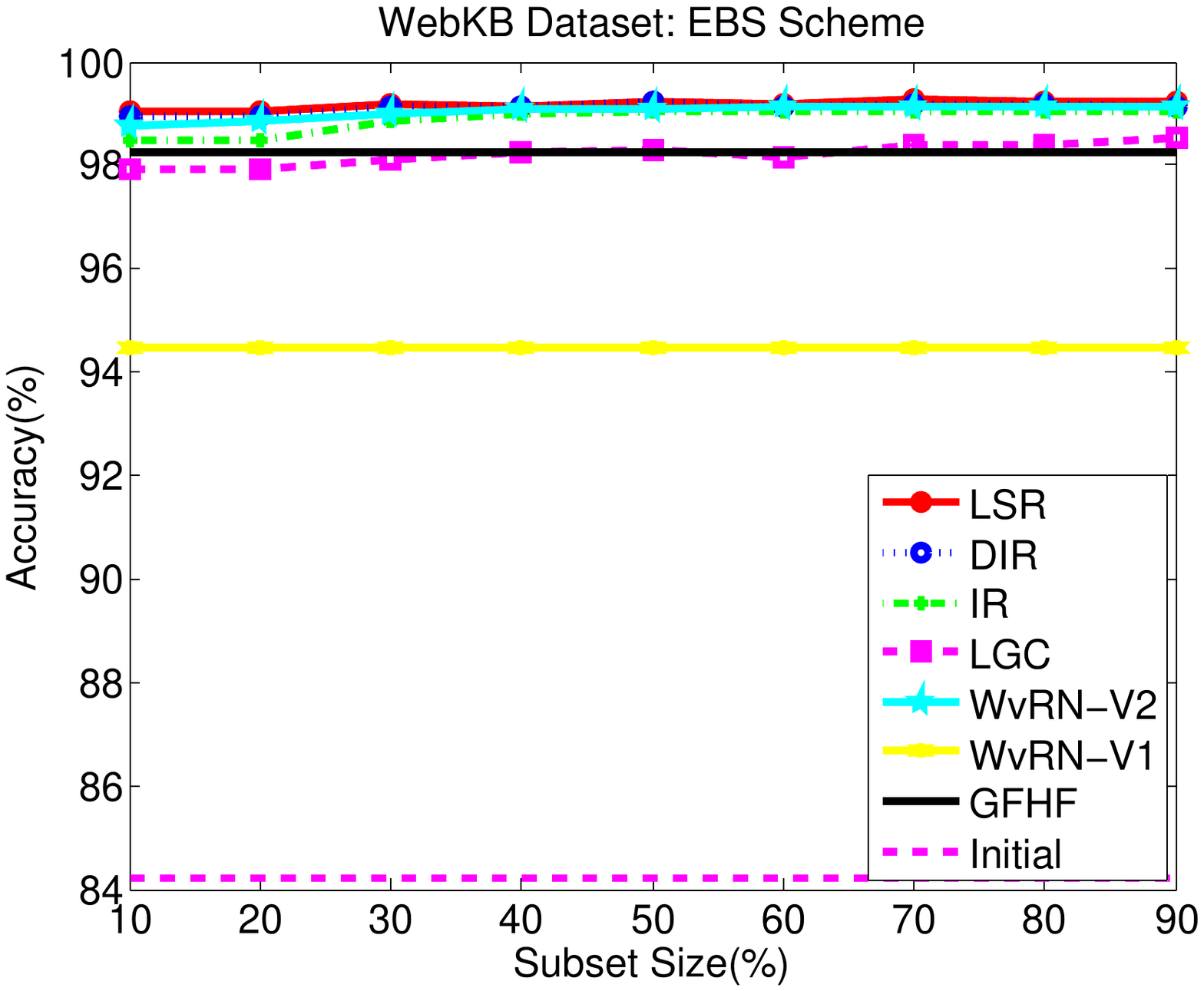} \\
\vspace*{-26ex}
		\includegraphics[width=0.4\linewidth]{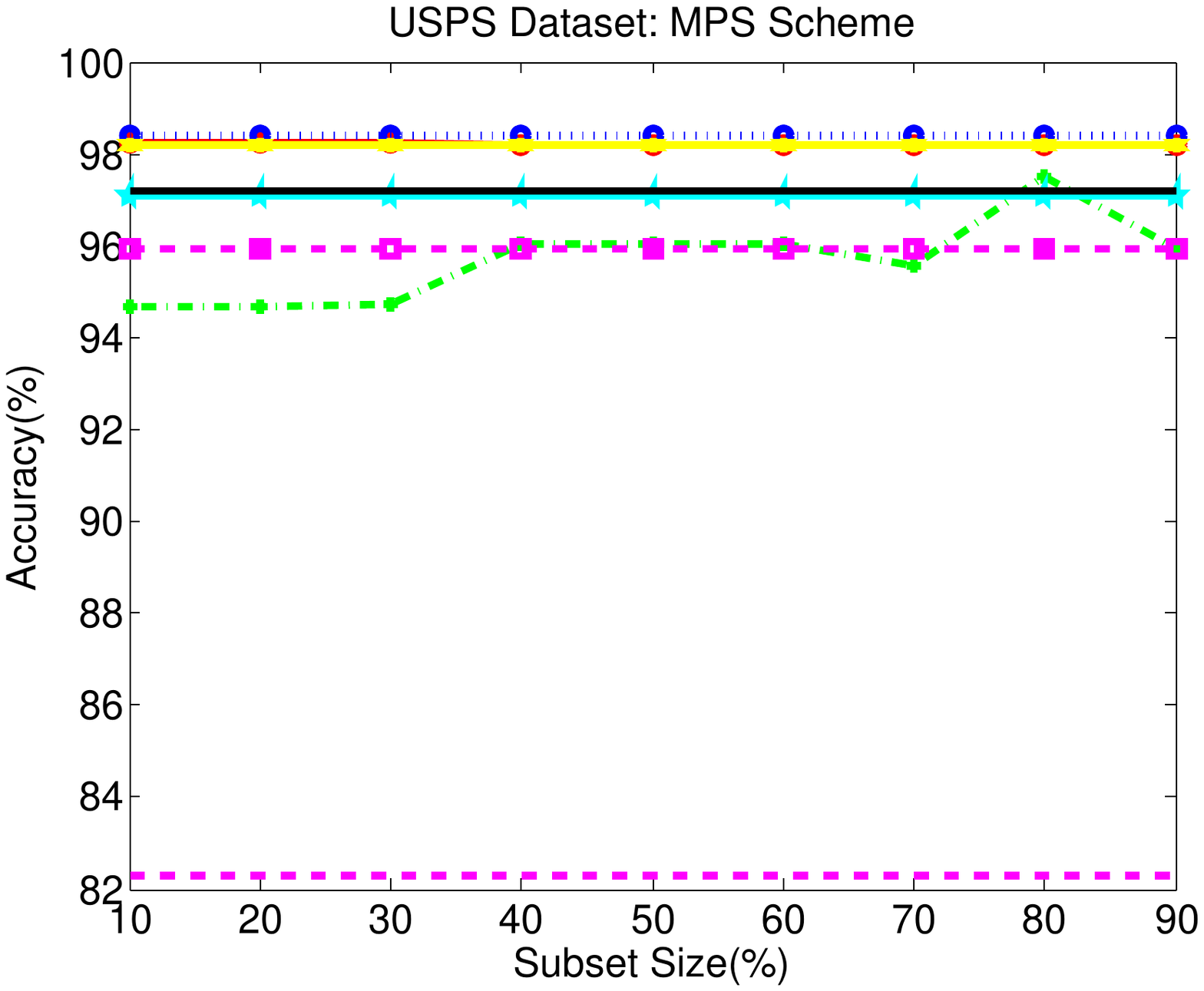}
		\includegraphics[width=0.4\linewidth]{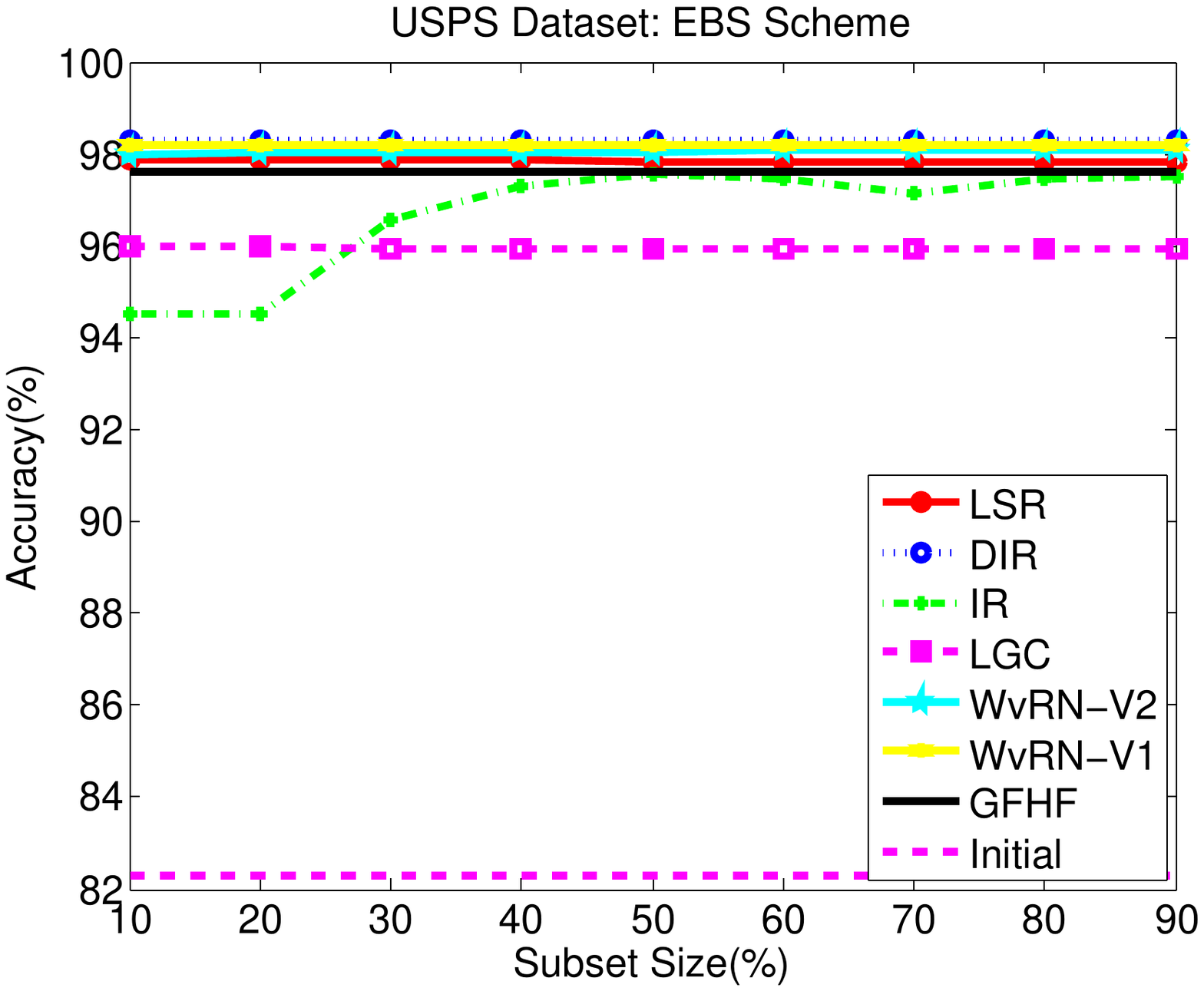} \\
\vspace*{-26ex}
		\includegraphics[width=0.4\linewidth]{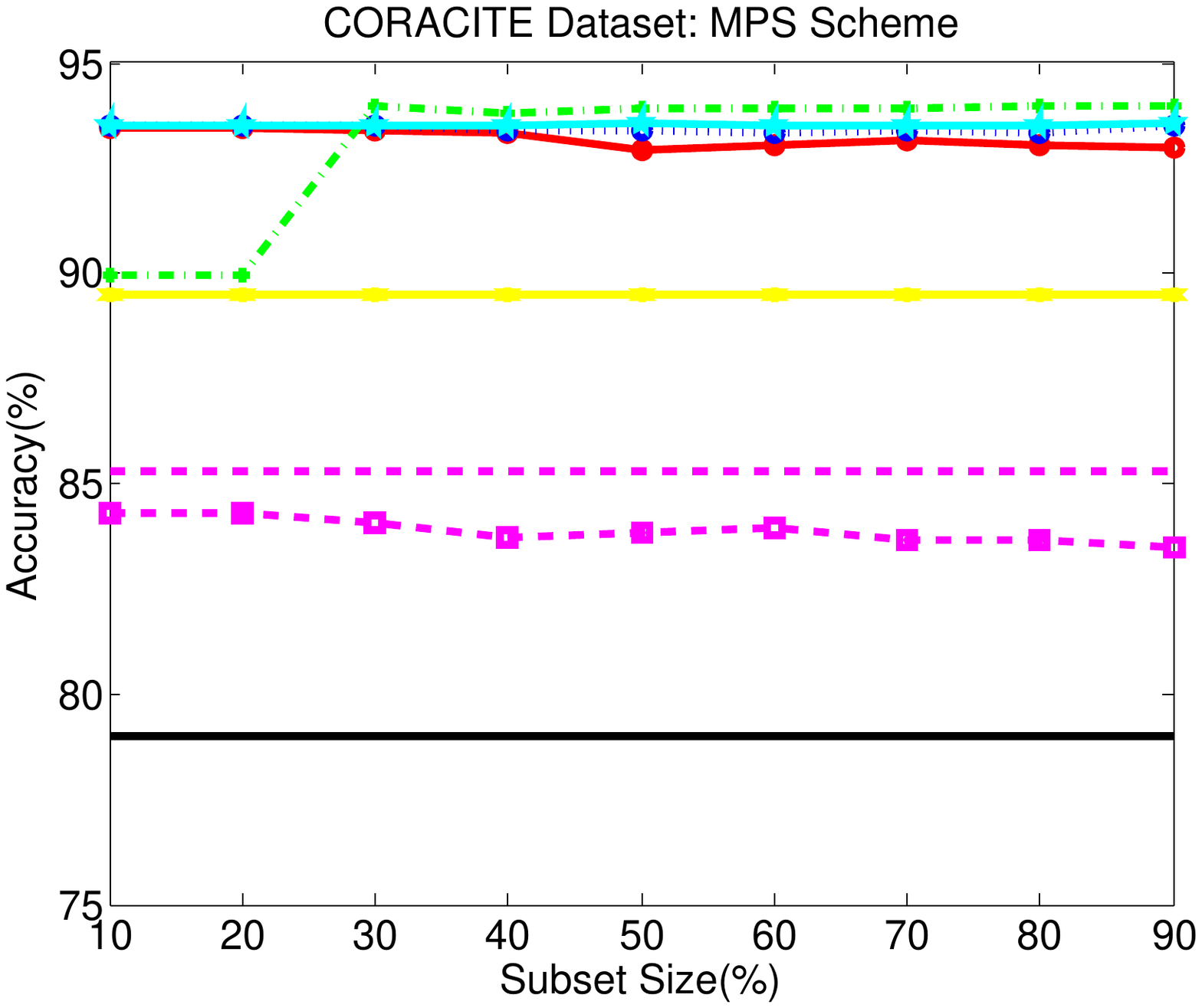}
		\includegraphics[width=0.4\linewidth]{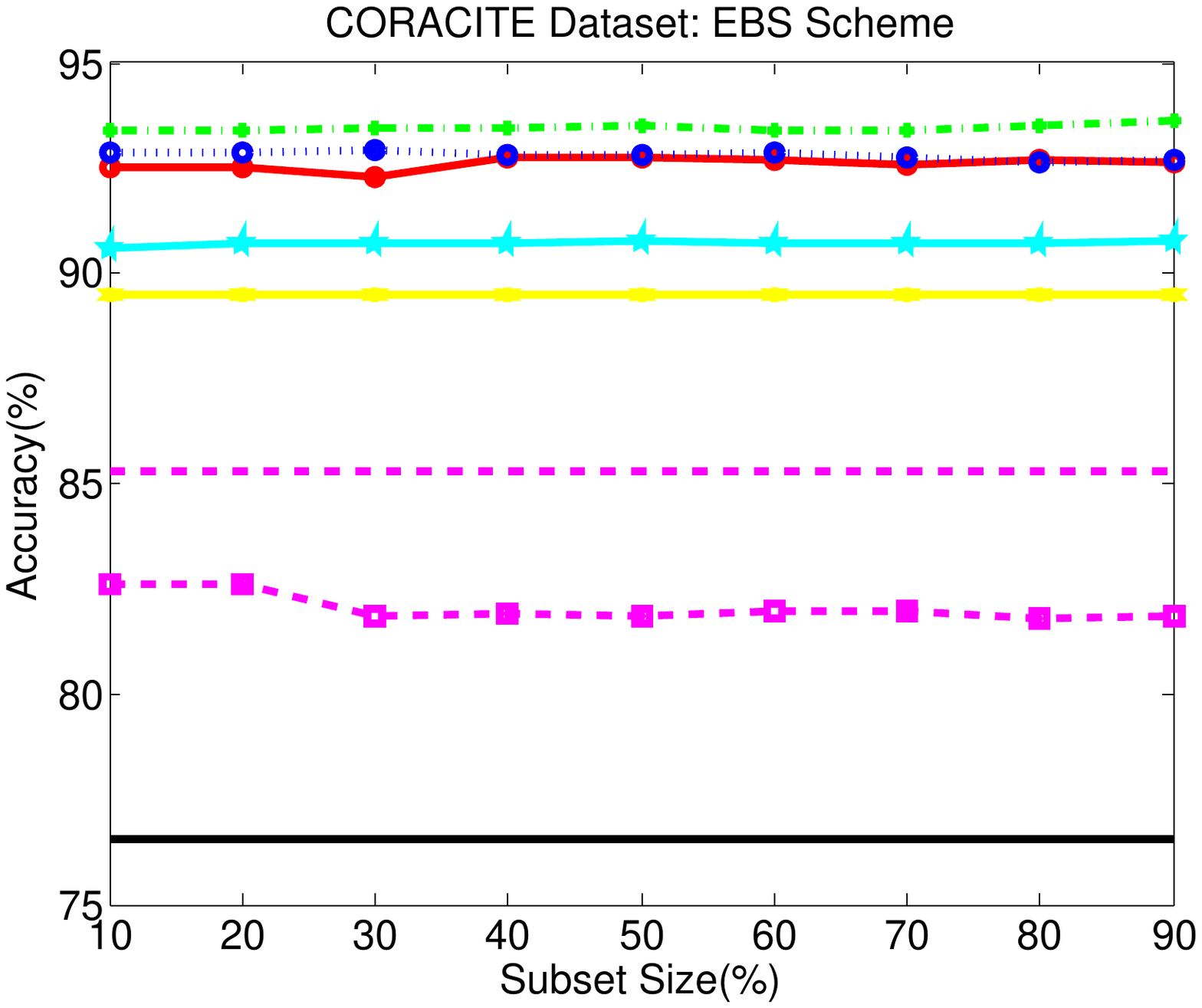} \\
\vspace*{-26ex}
		\includegraphics[width=0.4\linewidth]{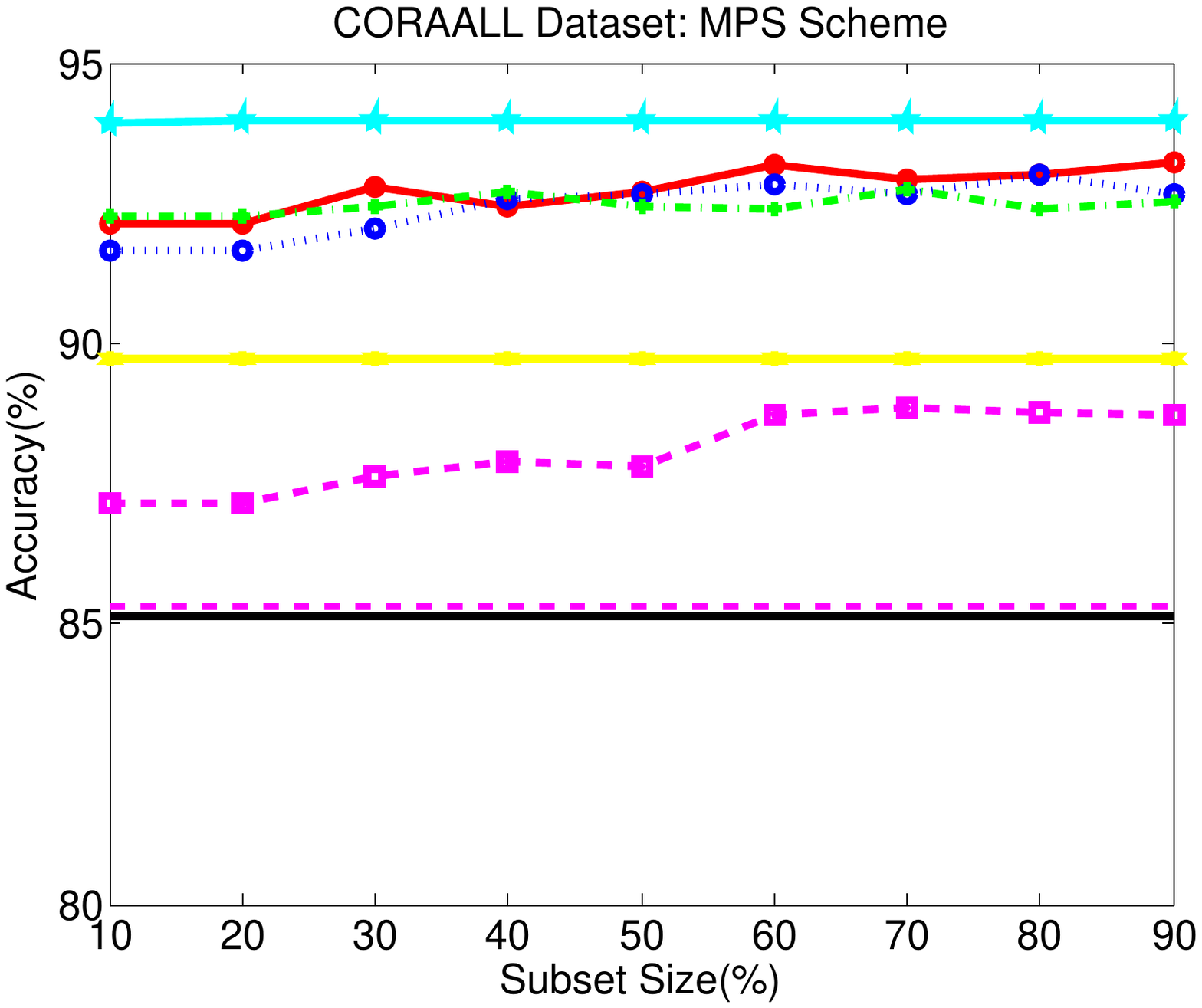}
		\includegraphics[width=0.4\linewidth]{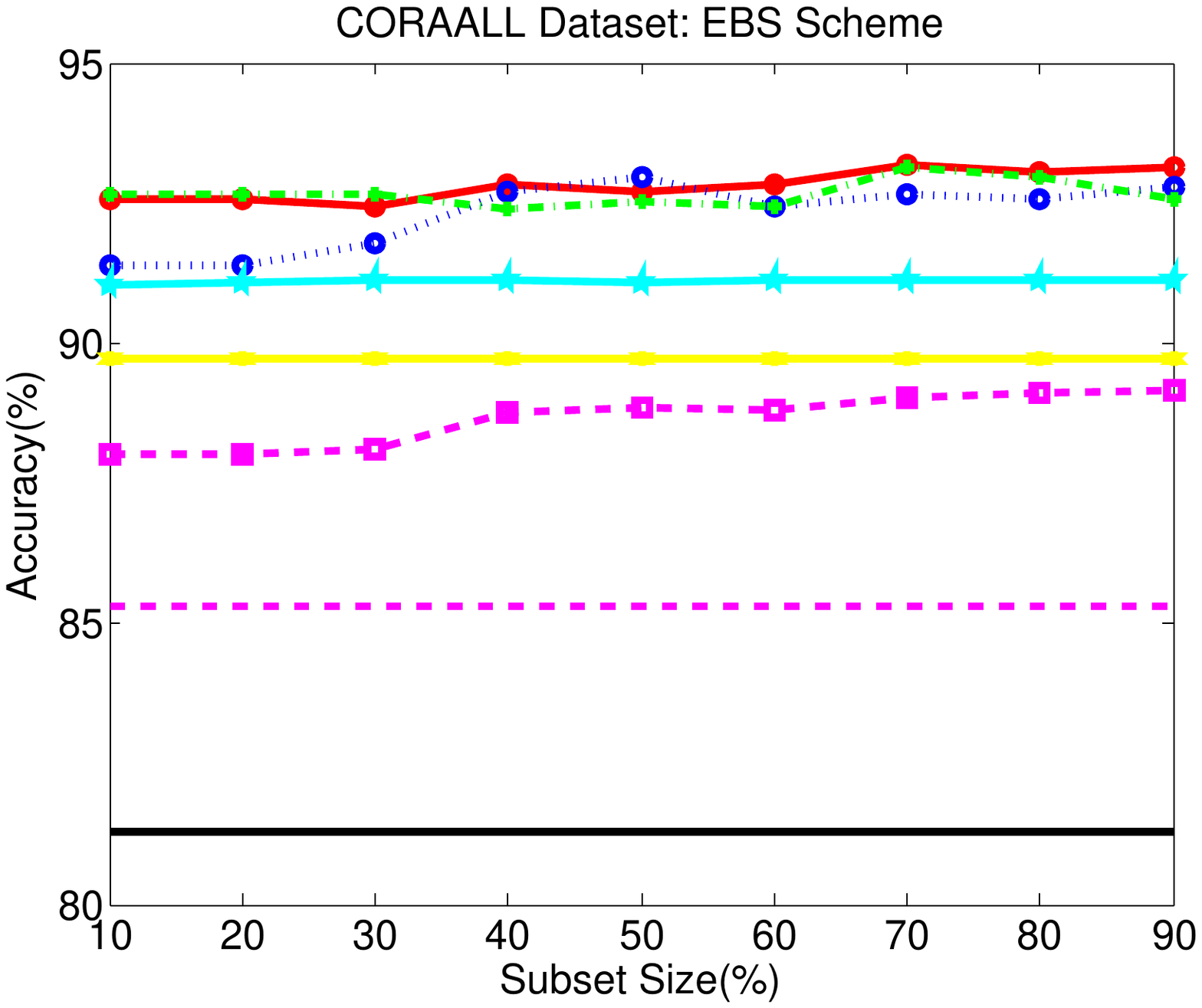} \\

\vspace*{-17ex}
 \caption{{Setting 2 Results: X-axis represents the subset size in terms of percentage of number of nodes and Y-axis represents accuracy. The left and right columns correspond to MPS and EBS subset selection schemes respectively. The legends for the CORA datasets are same as the one given for the WebKB and USPS datasets. {\it Initial} refers to accuracy computed from generated inaccurate distribution.}}
\label{fig:s2}
\end{figure*}

\subsection{Experiment With Setting 1.}
We conducted the experiment with the two function estimation methods, LGC and GFHF, and the four probabilistic methods, WvRN, IR, DIR and LSR.
We measured the classifier accuracy on the entire set of nodes. This experiment was conducted with both maximum probability scoring (MPS) and entropy based scoring (EBS) subset selection schemes. The selected subset size $|S|$ was varied from {\it top} 10 percent to 90 percent (as per the chosen selection scheme).

Let us discuss parameter selection.
Recall that in the GFHF method $C$ is set to 1. In the IR, DIR and LSR methods only the second term is present (once we fix the probability distribution of {\it labeled} nodes). In the WvRN method we set $\nu=0.95$. Thus only the LGC method required tuning of the regularization parameter $C$. For selecting $C$ we used 5-fold CV on the labeled nodes and varied $C$ in the range $[0.00153,100]$ (doubling in each step). As mentioned in section 3.4.3, note that the 5-fold evaluation used here is inaccurate since it is based on noisy information. Therefore, the $C$ estimate can be inferior sometimes.

The average classifier accuracy was computed as the average of accuracies obtained from 100 random partitions of the graph. The results for one parameter setting with the MPS subset selection scheme for all the datasets are given in Figure~\ref{fig:s1}; we set $p_{max}=0.99$ for all the datasets and set $p_{min}$ as 0.4, 0.2 and 0.1 for WebKB, USPS and CORA datasets (both CORACITE and CORAALL)  respectively. 

From Figure~\ref{fig:s1} we see that only on the WebKB dataset all the methods are able to perform better than the initial accuracy; in fact, even on this dataset, WvRN and GFHF methods fall short when the subset size $|S|$ is small. All methods other than LGC give almost the same accuracy as $|S|$ becomes larger, on all datasets; of course, the exact value of $|S|$ at which they converge vary from one dataset to another. On the CORA datasets WvRN performed better followed by GFHF, LSR, DIR, IR and LGC when $|S|$ is small. On comparing EBS and MPS selection schemes we found that the EBS scheme performed slightly better (around 2\%) on CORA datasets, while the performances were almost same on WebKB dataset; the EBS scheme performed slightly inferior on USPS dataset. Except for these variations, the behavior of the EBS scheme as a function of $|S|$ was almost same as that of the MPS scheme; therefore, only the results for the EBS scheme are shown in the figure. We analyzed the inferior performance of LGC on CORA datasets and observed that incorrect choice of $C$ due to noisy CV estimates was the reason behind it.

\subsection{Experiment With Setting 2.}
We conducted the experiment with LGC, GFHF, IR, DIR, LSR, WvRN-V1 and WvRN-V2. The classifier accuracy measurement, evaluation with two selection schemes, subset size variation and $p_{min}$ and $p_{max}$ settings remain the same as in setting 1.
In this setting IR, DIR, LSR and WvRN-V2 {\it also} require parameter tuning (that is, $C$ and $\nu$). In the case of WvRN-V1 there is no parameter tuning and we set $\nu=0.95$.
In the case of GFHF we set $C=1$ as done earlier. All parameter tunings are done using 5-fold CV, using top $M$ percent of selected nodes, as in setting 1. It is useful to recall that the main difference of setting 2 from setting 1 is that the distribution information from all nodes is used during the solution process. We used the same range for C as in setting 1 for the LGC method. In the case of WvRN-V2 method we used the same range but set $\nu=\frac{1}{1+C}$. In the modified IR method we selected $C$ from the range $[0.0625,312.5]$ (doubling in each step). In the case of LSR and DIR methods we selected $C$ from the range $[0.078,10]$ (doubling in each step).
Finally the average accuracy was computed from 100 random partitions of the graph for all the methods except IR.

{\bf Efficiency} The objective function optimization in the case of IR takes significantly longer time (an order of magnitude) compared to other methods in this setting due to the nonlinear optimization involved. Therefore, only for IR, we computed the average performance from 100, 20 and 10 partitions for USPS, WebKB and CORA datasets respectively. In terms of speed, WvRN, LGC, GFHF, LSR, DIR and IR methods come in that order. Although the update equations for the DIR and LSR methods look similar, the DIR method took more time to converge than the LSR method and we observed that the LSR method was approximately 6-10 times faster than the DIR method. On the other hand, the LSR method was slower by 2-4 times compared to the remaining methods.

{\bf Choice of C from Cross-Validation}
The choice of $C$ from the noisy CV accuracy estimates has an important effect on the classification performances of LGC, IR, DIR and LSR methods; this effect varies across datasets. In these methods we observed that inferior performances were obtained when we chose $C$ that gave the best CV accuracy. To get improved and robust performances, we recommend an alternate way to choose $C$: choose the smallest $C$ such that the corresponding CV accuracy estimate was within, say, 5$\%$ of the best CV accuracy. This helps because the variation of CV accuracy estimate as a function of $C$ can be quite flat around the best CV accuracy and, in such cases choosing the least $C$ within a specified accuracy estimate regularizes the solution better. The results reported in Figure~2 are the improved performances obtained using the recommended way of choosing $C$ described above; some inferior performances can be still seen with IR on the WebKB, USPS datasets (particularly for low values of $|{\bf S}|$) and LGC on CORA datasets. From this viewpoint, the results suggest that DIR, LSR and WvRN-V2 are more robust compared to IR and LGC.

{\bf Classifier Performance}
From figure~2 we clearly see that WvRN-V1, WvRN-V2, IR, DIR and LSR methods improved the performance significantly over the initial accuracy on all datasets. The LGC-extension and modified GFHF methods improved the performance significantly on WebKB and USPS datasets. The performance curves of GFHF and WvRN-V1 remain flat in each plot. This is because there is no parameter tuning involved in these methods. For the other methods the choice of $C$ made remained almost same for all $|{\bf S}|$. Considering the performance on all the datasets the EBS scheme seems to be more robust compared to the MPS scheme, although slightly inferior performance is seen on the CORAALL dataset. Note that the performance variations over $|{\bf S}|$ is lesser with the EBS scheme particularly on the WebKB dataset. We believe this behavior is due to the conservative estimate of $\lambda_{ii}$ given by this scheme compared to the MPS scheme in the noisy scenario. Clearly, compared to the first setting, the second setting using distribution information of all nodes during the solution process enhances the performance significantly.
Though function estimation methods performed quite close on two datasets, the probabilistic methods performed better considering all the datasets. Among the probabilistic methods, the performances of IR, DIR and LSR were quite close as $|{\bf S}|$ became large. Comparing WvRN-V2 and WvRN-V1, since the performance difference is significant ($>$4\% in many cases), WvRN-V2 is to be preferred. Since the performance variation across C in the WvRN-V2 method is not much it seems that much of the gain comes from using dongle nodes. Recall that GFHF and WvRN methods have almost same performances on all datasets in the first setting; however, in setting 2, distinctly different performances of these methods are seen, particularly on CORA datasets. More investigation is needed to understand this behavior.
Finally, the performance improvement is dependent on the quality of relational graph (with respect to the assumption of strong connectivity of nodes belonging to same class) and initial accuracy.

\begin{table*}
\begin{center}
\caption{\small{Experiment: Shopping Datasets. $n$, $L$ and $|E|$ denote the number of nodes, labeled nodes and edges in the graph respectively. The classificiation accuracy evaluated on the labeled nodes for the external classifier (EC),
LGC, WvRN-V2, LSR and DIR methods are given.}}
\vskip 0.05in
\begin{small}
\begin{tabular}{|l|c|c|c|c|c|c|c|c|} \hline
Problem & $n$ & $L$ & $|E|$ & EC & LGC-Ext & WvRN-V2 & LSR & DIR\\ \hline
CU-D & 53023 &1433 & 1186190  & 77.18 &77.88 &77.74 & {\bf 78.02} & {\bf 78.02} \\ \hline
CU-L & 53023 &1433 & 1186190 & 91.70 &93.79 &93.93 & {\bf 94.21} & {\bf 94.21} \\ \hline
UG-D & 82027 &1166 & 1714228 & 94.51 & {\bf 97.68} &94.51 & 95.54 & 95.80 \\ \hline
UG-L & 82027 &1166 & 1714228 & 81.90 & {\bf 91.51} &85.93 & 91.25 & {\bf 91.51} \\ \hline
WM-D &67997 &1250 & 1318903 & 93.20 &94.48 &95.12 & 95.52 & {\bf 95.60} \\ \hline
WM-L &67997 &1250 & 1318903 & 93.36 &94.72 &94.72  & {\bf 95.60} & 95.04 \\ \hline
\end{tabular}
\end{small}
\end{center}
\label{tbl2}
\end{table*}

\subsection{Experiment with Shopping Datasets.}
We evaluated the performances of DIR, LGC, WvRN-V2 and LSR in solution setting 2 on real world datasets from three shopping sites ({\it www.compusa.com}, {\it www.uncommongoods.com} and {\it www.walmart.com}).
We considered two binary classification problems: {\it product detail vs non-product} and {\it product listing vs non-listing}. While the product pages are about one specific product like canon camera with certain model number in more detail, the listing pages are about several products of same cateogory (for example, different models of canon camera) arranged as a list in each page. These problems along with their site names are referred to as CU-D, CU-L, UG-D, UG-L, WM-D and WM-L respectively in Table~2. In each problem, the class distribution score for each page was obtained using an external classifier (EC) based on content features. The relational graph was constructed using structural signature (shingle) obtained using html tags of web pages. An edge between two pages was formed when their structural signatures had a match score of at least 6 (the values are in the range [0,8]) and a unit weight was assigned to each such edge. Also, each node was connected to a maximum of 20 other nodes. A subset of pages (nodes) in each site (graph) was manually labeled; the number of labeled nodes is indicated as  $L$ in Table~2. The classifier accuracy was evaluated for the external classifier and the various methods on these labeled nodes. The results are given in Table~2.

In almost all the cases the EBS scheme performed better compared to the MPS scheme. The results given in Table~2 are the best performances obtained over the percentage of subset sizes ($|{\bf S}|$) and $C$ values. There were minor variations ($0.5-2\%$) in the performances over ($|{\bf S}|$) and choices of $C$. The results clearly demonstrate that significant performance improvement (as high as 10\%) can be achieved. Also, the usefulness of the proposed extensions and LSR method as effective alternate methods can be seen from comparison with the DIR method.

\section{Summary}
We considered the problem of collectively classifying entities where relational information and inaccurate class distribution information from another (external) classifier are available. We present below a list of key observations from the experimental studies conducted on several benchmark and real world datasets.
\begin{itemize}
\item Of the two solution settings evaluated, the second setting (which uses external classifier information of all nodes) is better. Using this second setting a significant improvement over the external classifier performance can be achieved using relational information.
\item For parameter selection, the entropy based selection scheme was observed to be more robust compared to the maximum probability selection scheme.
\item With respect to choice of $C$ using inaccurate CV estimates, the DIR, LSR and WvRN-V2 methods were observed to be more robust.
\item Overall, the probabilistic methods fared better compared to the function estimation methods. Within the probabilistic methods, the IR methods were competitive to the other methods. Within the proposed methods, the LSR, WvRN-V2 and LGC ranked better, in that order.
\item In terms of speed, the original IR method was quite slow compared to other methods. Although the DIR method was faster, it was still not competitive in speed to the proposed methods. Among the proposed methods, the WvRN-extensions, LGC and LSR methods were faster, ranked by speed in that order.
\item Overall, considering the issues of speed, robustness and accuracy together, the LSR and WvRN-V2 methods performed the best.
\end{itemize}


%
\scriptsize{
\bibliographystyle{abbrv}
\bibliography{gbmueci}}  

\begin{thebibliography}{1}

\bibitem{bengio}
Y.~Bengio, O.~Delalleau, and N.~L. Roux.
\newblock Label propagation and quadratic criterion.
\newblock In O.~Chapelle, B.~Sch{\"o}lkopf, and A.~Zien, editors, {\em Label
  Propagation and Quadratic Criterion}. MIT Press, 2006.

\bibitem{cord}
A.~Corduneanu and T.~Jaakkola.
\newblock Distributed information regularization on graphs.
\newblock In {\em NIPS}, pages 297--304, 2005.

\bibitem{klein}
J.~Kleinberg and E.~Tardos.
\newblock Approximation algorithms for classification problems with pairwise
  relations: Metric labeling and markov random fields.
\newblock In {\em FOCS}, pages 14--23, 1999.

\bibitem{sofus}
S.~A. Mackassy and F.~Provost.
\newblock Classification in networked data: A toolkit and a univariate case
  study.
\newblock {\em JMLR}, 8:935--983, 2007.

\bibitem{priti}
P.~Sen, G.~M. Namata, M.~Bilgic, L.~Getoor, B.~Gallagher, and T.~Eliassi-Rad.
\newblock Collective classification in network data.
\newblock Technical Report CS-TR-4905, University of Maryland, 2008.

\bibitem{koji}
K.~Tsuda.
\newblock Propagating distributions on a hypergraph by dual information
  regularization.
\newblock In {\em ICML}, 2005.

\bibitem{wang}
J.~Wang, T.~Jebara, and S.-F. Chang.
\newblock Graph transduction via alternating minimization.
\newblock In {\em ICML}, 2008.

\bibitem{zhou}
D.~Zhou, O.~Bousquet, T.~Lal, J.~Weston, and B.~Scholkopf.
\newblock Learning with local and global consistency.
\newblock In {\em NIPS}, pages 321--328, 2004.

\bibitem{zhu}
X.~Zhu, Z.~Ghahramani, and J.~Lafferty.
\newblock Semi-supervised learning using gaussian fields and harmonic function.
\newblock In {\em ICML}, pages 912--919, 2003.

\end{thebibliography}
%
%

\end{document}